\RequirePackage{fix-cm}

\documentclass[twocolumn]{svjour3}          
\smartqed  
\usepackage{graphicx}
\usepackage[numbers]{natbib}
\usepackage[colorlinks,citecolor=blue,urlcolor=blue,bookmarks=false,hypertexnames=true,pagebackref=false]{hyperref}

 \usepackage{mathptmx}      
 \journalname{International Journal of Social Robotics}

\begin{document}

\title{Identifying Functions and Behaviours of Social Robots during Learning Activities
}
\subtitle{Teachers' Perspective}

\author{Jessy Ceha         \and
        Edith Law \and
        Dana Kuli\'c \and
        Pierre-Yves Oudeyer \and
        Didier Roy
}

\institute{J. Ceha \and E. Law  \at
              University of Waterloo, 200 University Avenue West, \\Waterloo, Ontario, N2L 3G1 \\
              \email{jceha@uwaterloo.ca}         
              \and
            D. Kuli\'c \at
            Monash University, Victoria 3800, Australia
            \and
            P-Y. Oudeyer \and D. Roy \at
            Centre Inria Bordeaux - Sud-Ouest, 200 avenue de la Vieille Tour, 33405, Talence, France
}

\date{Received: date / Accepted: date \textcolor{red}{This is a preprint (before peer-review) of an article published in The International Journal of Social Robotics. The final authenticated version is available online at: https://doi.org/10.1007/s12369-021-00820-7}}

\maketitle

\begin{abstract}
With advances in artificial intelligence, research is increasingly exploring the potential functions that social robots can play in education. As teachers are a critical stakeholder in the use and application of educational technologies, we conducted a study to understand teachers' perspectives on how a social robot could support a variety of learning activities in the classroom. Through interviews, robot puppeteering, and group brainstorming sessions with five elementary and middle school teachers, we take a socio-technical perspective to conceptualize possible robot functions and behaviours, and the effects they may have on the current way learning activities are designed, planned, and executed. Using activity theory to analyze learning activities as an activity system illustrated a number of tensions that currently exist between the components of the system. Overall, the teachers responded positively to the idea of introducing a social robot as a technological tool for learning activities, envisioning differences in usage for teacher-robot and student-robot interactions. We discuss the fine-grained functions and behaviours envisioned by teachers, and how they address the current tensions - providing suggestions for improving the design of social robots for learning activities. 

\end{abstract}

\section{Introduction}
\label{intro}

Advances in artificial intelligence (AI) are enabling new applications in educational technology, such as robots \citep{belpaeme2018social,Gaudiello16}.  Robots have been used in the classroom as a tool for explaining science, technology, engineering, and mathematics (STEM) concepts (for reviews see: Benitti \cite{benitti2012exploring}; Mikropoulos \& Bellou \cite{mikropoulos2013educational}), to provide remote access (e.g., \cite{cha17}), and as an embodied social agent that acts as a teacher or classroom companion (e.g., \cite{Kanda2004}).  Our focus is on exploring how robots, as social agents, should interact with teachers and students in the classroom, specifically during learning activities.  This question has been partially answered by prior works investigating field deployment of social robots in classroom settings (e.g., \cite{Kanda2004,Tanaka2012}), and explorations of the design space for robot roles through consultation with teachers (e.g., \cite{Ahmad2016,Westlund2016}).  In most of these prior works, educational robots are thought to have three main roles---teacher, peer, and novice. However, beyond these roles, little is known about teachers' perspectives on what exactly social robots should {\it do} in these roles, and {\it why}.

Our work aims to fill this gap by using activity theory (a framework for analyzing artifact-mediated and objective-driven human activity \cite{Engestrom2010,nardi09,Kaptelinin2012}) and participatory design methodologies to analyze robot roles during learning activities, and consider the finer-grained {\it functions and behaviours}. A study was conducted with five elementary and middle school teachers over several months, which included semi-structured interviews, robot puppeteering, and group brainstorming, to understand how they design, plan, and execute learning activities, and how they envision social robots can participate.  We analyze the qualitative data by framing learning activities as an activity system, analyze the tensions within this system, and identify robot functions that help alleviate these tensions.  

In summary, this work contributes an analysis of learning activities as an activity system, a conceptualization of robot roles as functions and behaviours that help alleviate tensions in the activity system, and considerations for future directions.

\section{Related Work}
\label{sec:1}
\subsection{Social Robots in Education}
\label{sec:2}
The use of social robots -- robots designed to interact and communicate with people -- in educational settings is an area of growing research interest. In contrast to other educational robots, for example those used as tools for teaching programming skills, social robots teach through social interaction \cite{belpaeme2018social}. The physical presence of robots can elicit higher rates of compliance to requests compared to the same robot displayed on a video \cite{bainbridge2011}, and compared to voice-only and virtual agents, students show more interest and higher performance while learning using physical robots \cite{Leyzberg2012}.

\subsection{Roles of Robots}
\label{sec:3}
According to several comprehensive reviews \cite{belpaeme2018social,mubin13}, social robots for education are found to most frequently take the role of a \textbf{teacher/tutor} (e.g., \cite{Belpaeme2018Tutor, you06, Alemi2014, Hashimoto2013, Movellan2009, Kanda2004, Kanda2007, Hyun2008}), \textbf{peer/co-learner} (e.g., \cite{Kanda2004,zaga15,baxter17,Lubold2016}), and \textbf{novice} (e.g., \cite{Chase2009,tanaka07,Tanaka2012,Hood2015}). The teacher/tutor role is usually an assistant to the human teacher, peers/co-learners are based on approaches of cooperative learning and can lead to collaboration between student and robot, and the novice allows the student to act as the tutor and teach the robot. In these roles, researchers have envisioned social robots that offer a variety of behaviours for instructional support, such as
telling stories to engage preschool children in constructive learning \cite{fridin14}, tailoring content to students based on performance \cite{Leyzberg2014,Gordon2015}, guiding attention through social dialog for vocabulary training \cite{Saerbeck2010}, personalizing support to promote intrinsic motivation \cite{Janssen2011}, and displaying empathy \cite{Leite2014}. 
Robots for learning are commonly developed to impact cognitive outcomes such as knowledge acquisition and comprehension (measured as amount and accessibility of knowledge or accuracy of recall) and affective outcomes such as attitudes toward learning and perceived performance capability (measured by self-reports or observations by experimenters).

Prior works commonly investigate the utility of social robots in learning environments using lab experiments and in a largely techno-centric perspective. Some researchers, however, have suggested the importance of involving stakeholders during the design process (e.g., \cite{vsabanovic2010robots}). Teachers are an especially critical stakeholder as they determine how and when children use technology, will be required to work with and integrate the robots into their classrooms, and cope with changes the introduction of this new technology may have on their established educational practices \cite{nordkvelle2005visions}.

\subsection{The Teacher's Perspective}
\label{sec:4}

There have been several prior studies examining teachers' perception and attitude towards educational robots \cite{serholt2014teachers,Broadbent2018,Serholt2017,van2020teachers,Westlund2016,Ahmad2016,chang10}. Some have considered teachers' perspectives on the moral and ethical implications of social robots. Serholt et al. \cite{Serholt2017} investigated this through a series of focus groups with 77 pre-service and primary school teachers from Sweden, Portugal, and the U.K. The teachers expressed concerns about privacy of students' data, a fear of losing boundaries between teacher and robot roles/functions/tasks, the potential for interactions with a classroom robot negatively impacting the students, and confusion over responsibility for the robot. Van Ewijk, Smakman, and Konijn \cite{van2020teachers} conducted focus group sessions with eighteen teachers from multiple schools in The Netherlands. The NAO robot was used as an example of a social robot and, similar to Serholt et al. \cite{Serholt2017}, the researchers found that the moral values considered most relevant to social robots in education by the teachers were related to privacy/security, psychological welfare/happiness, applicability, and usability. 

Other studies have focused more on the roles and applications of social robots in the classroom. Serholt et al. \cite{serholt2014teachers} interviewed eight teachers across England, Scotland, Portugal, and Sweden and found concerns related to robots being disruptive, the robot needing to be able to manage group work, and a fear of robots replacing human interaction. However, the teachers also felt that a robot could be beneficial by reducing their workload in terms of teaching and assessment, and that they could be used to guide, motivate, and encourage students. Ahmad, Mubin, and Orlando \cite{Ahmad2016} introduced eight language teachers from primary and high schools to the NAO robot and conducted an interview to solicit their opinions on how the robot could contribute to language learning. Common roles and behaviours the teachers mentioned were to motivate and encourage the children, to adapt behaviour to different scenarios, and to have a sense of humour. Broadbent et al. \cite{Broadbent2018} had 207 students and 22 teachers in New Zealand interact in 30 minute sessions with Paro and iRobiQ (two popular companion robots) to explore whether they may be useful in rural schools. Attitudes were generally positive, especially those expressed by children aged 5-12 and their teachers, and the functions thought to be most useful included helping children with autism, comforting children in sick bay, and for repeating exercises to children who need extra help. 

Researchers have also deployed robots in classrooms to investigate teachers' perspectives. Kory-Westlund et al. \cite{Westlund2016} interviewed thirteen teachers and teaching assistants in three American public schools about their views prior to and following deployment of robots in their preschool (3-5 years) classrooms. The teachers were mostly positive about the use of robots in the classroom and although they expected the robots to cause a disruption, this was not the case. A common suggestion from teachers was to use the robot to teach social skills such as turn-taking, patience, and sharing to the children. Another study looking at the use of a social robot for language learning, deployed a humanoid robot in three fifth-grade classrooms in Taiwan \cite{chang10}. A total of 100 students and three teachers interacted with the robot during five scenarios: storytelling, oral reading, cheerleader mode, action mode, and question-and-answer mode. Teacher and student reactions indicated the robot could be useful in creating an engaging and interactive environment. However, the teachers voiced three main concerns: cost, complexity of using robots for teaching, and requiring appropriate learning content to be used with the robot.

\subsection{Activity Theory}
\label{sec:5}
Activity theory is a framework for analyzing needs, tasks, and outcomes, providing a lens through which researchers can analyze human activity. It has been used in Human-Computer Interaction research, for example to look at the social media use of teens \cite{yardi11}, and how users on Wikipedia transform from novice to expert \cite{bryant05},
as well as in Human-Robot Interaction research for analyzing student behaviour during mathematics learning with robots \cite{fernandes2010robot}, as a basis for the design of a Robot Behaviour Toolkit \cite{huang2012robot}, and for structuralizing behaviours in a robot-assisted intervention for children with autism \cite{kim2014designing}. 

Engeström's Activity System Model \cite{engestrom1987}, a variant of activity theory, extends the individual subject-object interaction \cite{leontiev1978}, to a collective activity -- an activity system. It consists of six components that interact to produce an outcome, (a) subject: the person doing the deed, (b) object: the deed being done, (c) community: all the people who are involved in or have influence on the activity, (d) tool: tools that help to allow the subject accomplish the object, (e) rule: norms and conventions that govern the relationship between subject and community, and (f) division of labour: how the work is divided among community members to achieve the object. Engeström proposed that activity systems are continuously developing as a result of \textbf{contradictions} -- conflicts or tensions within and between activity systems -- and that these contradictions have an impact on the evolution of the activity system as a whole \cite{Engestrom2001}. Identifying contradictions in a system provides a way for studying how changes in any component of the system (e.g., the adoption of a new tool) may alleviate, worsen, or introduce tensions in the system. This approach has been used to study various aspects of technology use in educational contexts (for a review see: Murphy \& Rodriguez-Manzanares \cite{murphy2008using}). Although closely related models for analyzing human activities exist, including: Distributed Cognition \cite{Hollan2000}, Phenomenology \cite{Winograd1987}, and Situated Action \cite{Suchman1987}, in our work we use Engeström's Activity System Model, as the introduction of a robot to a classroom setting can be seen as changing a key component of an activity system.\\

\begin{figure}
    \centering
    \includegraphics[width=\columnwidth]{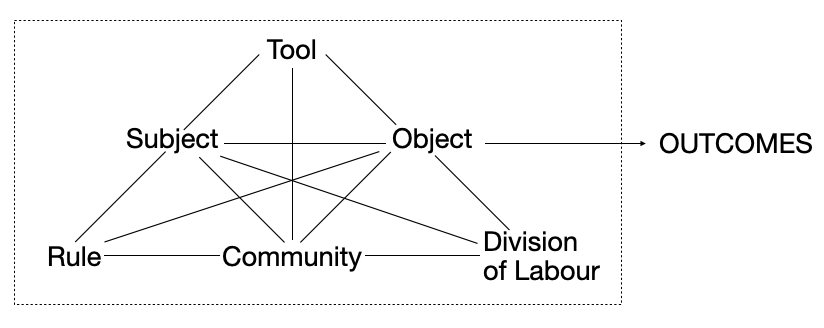}
    \caption{Engeström’s Activity System Model \cite{engestrom1987}.}
    \label{fig:my_label}
\end{figure}

\section{Methodology}
\label{sec:6}

Our work aims to examine the interconnections surrounding learning activities (time-limited, in-class activities aimed at teaching a particular concept) using Engeström's Activity System Model, to understand how teachers envision themselves and their students interacting with a social robot during learning activities, and how these robot behaviours may alleviate or exacerbate tensions in the system. To achieve this, we use semi-structured interviews, puppeteering, and group brainstorming to provide (1) a rich description of the socio-technical contexts surrounding learning activities in which robots can be embedded, (2) enactments of situations related to particular learning activities, and (3) conceptualizations of robot functions and behaviours that are considered by all teachers to be useful. Our methodology and focus contrasts and supplements prior work in that we use activity theory to analyze teachers' processes of designing, organizing, and executing learning activities. Further, the majority of studies in this area present teachers and students with a pre-existing robot system, rather than inviting them to co-design. Instead, we have teachers puppeteer the robot during role-playing, providing them with a way to more concretely conceive the possible scenarios in which a social robot could be applied (a limitation  noted  in  prior  work, e.g., \cite{van2020teachers,Ahmad2016}), and solicit their thoughts on the robot's finer-grained verbal behaviours (as suggested by Serholt et al. \cite{serholt2014teachers}).

\subsection{Participants}
\label{sec:7}
Five female teachers (T1 - T5) were recruited through the principal of a local private school in Canada (Kindergarten to Grade 8---students aged 4-14 years), and compensated \$60 for Session 1, and \$80 for Session 2. To protect participants' anonymity and privacy, we provide a summary of their characteristics. T1 - T3 are generalist teachers, who teach all subjects, T4 teaches art, and T5 teaches biology exclusively to Grades 7-8. Our teachers had a variable amount of teaching experience (1-25 years) and at the time of the study worked with students between Grades 4 and 8 (i.e., ages 9-14).

\subsection{Procedure}
\label{sec:8}
The study was conducted over two sessions. The first was a one-on-one session (focused on initial exploration and discovery) lasting 1.5 hours; the second was a group ideation session lasting 2 hours.  
\paragraph {Session 1} In the first session, two researchers began by conducting a 45-minute {\bf semi-structured interview} with each teacher separately, to learn about their current methods for preparing and executing learning activities---the challenges they face, strategies they employ, and factors they consider when implementing learning activities. To ground our conversation in real experiences, we asked the teachers to think of a concrete example of a recent or favorite learning activity, and bring along to the interview materials that they used to prepare the learning activity, as well as any materials they would distribute to their students during the learning activity.  

Each interview started with questions about background, such as how long they have been teaching, what subjects and grades they have taught and currently teach, and what kinds of technologies they have had students use in learning activities.  We then asked three types of questions related to: (1) {\it Process:} Teachers walked through and described their chosen learning activity, including targets and objectives, sequencing of activity components, materials used, whether students worked individually or in groups, and other constraints or considerations they had to take into account,  (2) {\it Design and Preparation:} As our goal was to understand how social robots can be incorporated into learning activities, we were also interested in how teachers design and prepare learning activities in the first place.  What resources were used in planning?  Did teachers re-use ready-made activities or adapt them in any way?  What kind of preparation was needed and how much time was required?,  (3) {\it Execution:} Finally, we were interested in the teacher's experience running the learning activity and their personal role in the learning process.  Was the learning activity directed or did it allow for exploration?  What was the criteria for success?  What were the challenges and what strategies did teachers employ to mitigate these challenges?  If teachers could have access to a teaching aide, what would they want this helper to do?  

\begin{figure}[htbp!]
    \centering
    \includegraphics[width=0.9\columnwidth]{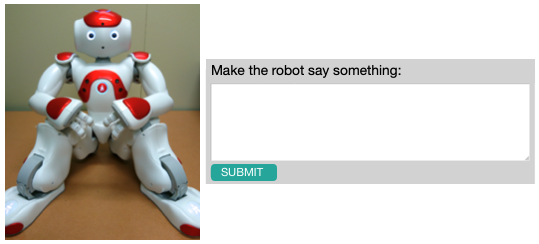}
    \caption{NAO robot and tele-operation interface used for puppeteering.}
    \label{fig:nao}
\end{figure}

During the second 45-minutes of the first session, teachers were introduced to NAO, a small humanoid robot developed by SoftBank Robotics, and given the opportunity to \textbf{puppeteer} the robot through a simple tele-operation interface that consisted of a text box for entering what they wanted the robot to say (Figure \ref{fig:nao}). The two interviewers and the teacher play acted the learning activity provided by the teacher using different scenarios, namely lecture-style (involving a teacher, a student, and the robot) and group-style (involving one or two students and the robot); each person took turns playing the role of the teacher, robot, or student.  Our goal here was to provide teachers with a platform to envision {\it specific} verbal behaviour that the robot could exhibit within the context of a particular learning activity, rather than merely talking about robot roles in general and hypothetical terms. After each scenario, we asked the teacher to explain why they had the robot exhibit the behaviours it did in the role-play.

\paragraph{Session 2} During the second session, we met with the five teachers at the same time to brainstorm specific functions and behaviours that the robot could employ during a learning activity---functions and behaviours that could be common across multiple disciplines.  We ran an activity called \textbf{635-brainwriting} \cite{rhorbach1969kreative}, where each person in a group of six, writes down three design ideas on a piece of paper in five minutes, then passes the sheet to the person next to them. The next person can modify/expand on the previous ideas, or write down new design ideas.  The process repeats until the sheet of paper returns to the original person. The design ideas were prompted by the question: \textit{What can the robot do to increase students’ interest in a topic?} Teachers were reminded what the general capabilities of the robot were (referencing the NAO robot), and not to worry about its potential limitations. 

All sessions were audio-taped, and photographs of relevant artifacts were captured.

\begin{figure*}[t!]
    \centering
    \includegraphics[width=0.99\textwidth]{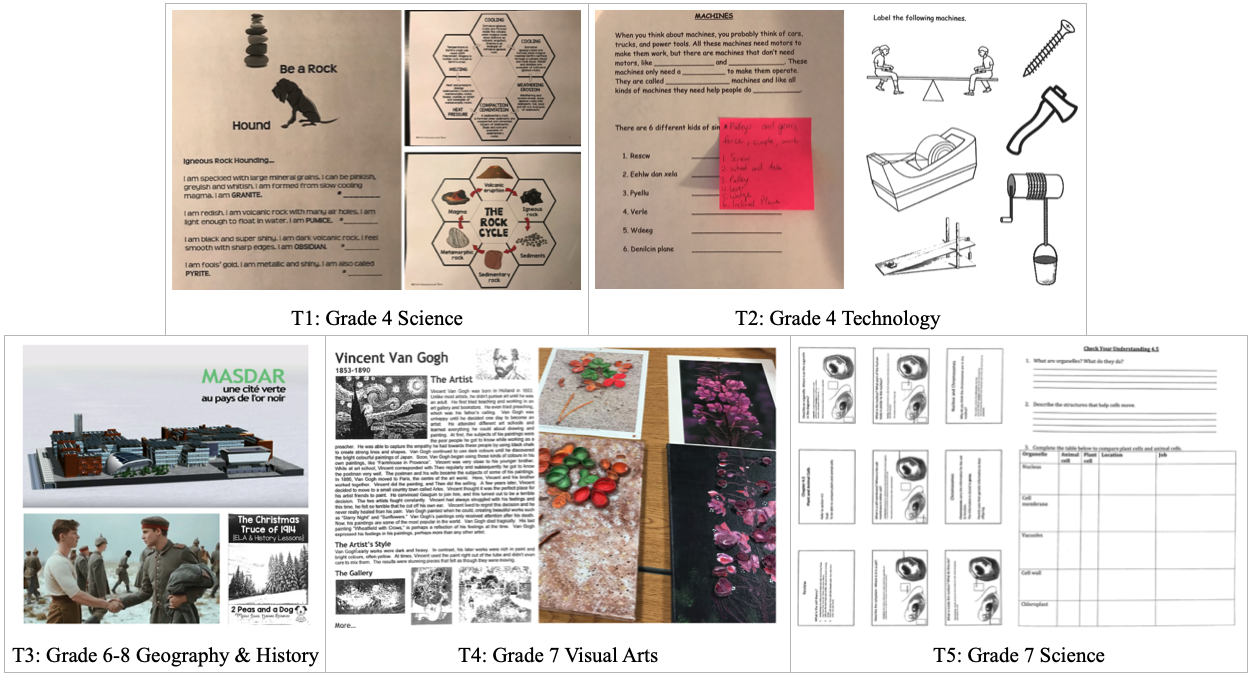}
    \caption{Teachers' Learning Activities}
    \label{fig:all_activities_2}
\end{figure*}

\subsection{Learning Activities}
\label{sec:9}
Despite our small sample, the learning activities discussed during the interviews were diverse in terms of domain, structure, and target age group - making the qualitative data a rich source of information about how teachers design and execute learning activities. The following is a brief description of the learning activities chosen by each interviewee:

T1: \textit{Rock Identification} aims to teach Grade 4 students how to classify rock types. The teacher uses two class times to cover the topic; in the first class explaining the learning objective, and then starting the activity involving the entire class by showing a video. Afterwards, they organize {\it `think-pair-share'}, where students work in pairs to complete a paper-based task about the rock cycle, then come together again to discuss what they learned. The class ends with a catchy ``rock song'' video.  In the second class, T1 splits the students into groups to complete a hands-on scavenger hunt, where students have to identify different rocks.

T2: \textit{Pulleys} aims to teach Grade 4 students what simple machines are (e.g., pulleys, incline planes), how they function, and what they are useful for.  The teacher takes a {\it `throw-them-in-the-dark'} approach, where pairs of students, without being taught anything about simple machines, are given fill-in-the-blank worksheets to complete without explicit guidance or instruction.

T3: Two activities for Grade 6-8 Geography and History.  For the Geography activity, \textit{Ideal Community}, students read a textbook chapter on communities before watching a video about Masdar city, a city built specifically to be sustainable.  Students are then asked to build their own ``ideal'' community, using their material of choice (e.g., Lego, paper, Plasticine).  In the History activity \textit{Christmas Truce}, students are shown the Sainsbury's commercial depicting the Christmas Truce of 1914, followed by an open-ended discussion about the video, where the teacher tries to draw a connection between the video and the students' lives. 

T4: \textit{Van Gogh Inspired Art} aims to teach Grade 7 students about elements of art (e.g., line, shape, colour, texture), principles of design (e.g., balance, contrast), and how to apply them in a painting.  The teacher starts by providing students with a handout about the life of Vincent van Gogh, shows them 5 images of van Gogh's famous paintings, and discusses their style.  Students are then asked to go home and research the artist using selected websites, and write down anything they find interesting about him. In the next class, the students look at the paintings on the SMART Board together. They are given a sheet listing the elements of art, and taught how to identify them in a painting. Finally, they are asked to create a realistic sketch of a photo of nature, given a rubric which evaluates not only their art, but also the quality of their exploration (i.e., research about the artist) and reflection (e.g., how they feel about the process versus the product, what parts of the painting they like/dislike and why).  

T5: \textit{Organelles and Cells} aims to teach Grade 7 students about the structure of cells.  The teacher shows a PowerPoint presentation about organelles and the differences between plant and animal cells.  Students then work in groups on a ``Check Your Understanding'' worksheet.  Finally, the teacher discusses the answers with the class and shows a video of how organelles move within the cell.

\subsection{Analysis}
\label{sec:10}
Two researchers transcribed the interviews and categorized the content by the six components of an activity system---Subject, Object, Tool, Rule, Community, and Division of Labour---through a thematic analysis using affinity diagramming. Based on this activity framework and the results gathered from the interview, brainstorming, and puppeteering sessions, we ask the following questions: 
\begin{enumerate}
    \item What tensions exist in the current activity system?
    \item How do teachers envision a social robot being incorporated?
    \item How may the introduction of a social robot alleviate or exacerbate the tensions?
\end{enumerate}  

\section{Results}
\label{sec:11}

\subsection{\textbf{RQ1: What tensions exist in the current activity system?}}
\label{sec:12}
Adopting the activity theory approach, in-class learning activities can be seen as an activity system.  Teachers are the {\bf subject} of the system; the {\bf object} is the learning activity.  Everyone that may be involved in the learning activities, e.g., other teachers, students, and parents, form the {\bf community}. {\bf Tools} help teachers design, plan, and execute learning activities---these can include technological as well as pedagogical methods (e.g., self-reflection). The use of technology by the teachers includes Chromebooks, SMART Board, YouTube, Google Classroom, PowerPoint, GoNoodle, Go-Animate for Schools, Kahoot, Pinterest, trusted online resources (e.g., the Canadian Encyclopedia), and an online marketplace for sharing teaching materials called TeachersPayTeachers.  {\bf Rule} specifies the constraints, norms, and conventions that govern how teachers should act within their community. For example, teachers must, for the most part, plan learning activities that follow the curriculum specified by the government, and work within the parameters of the particular school's schedule and resources. They also have to maintain fairness and equal opportunities for all students in the class.  Finally, {\bf division of labour} captures the sharing of responsibility between members of the community -- including students, parents, and other teachers -- in the design, planning, and execution of learning activities.  

\begin{figure*}[ht!]
    \centering
    \includegraphics[width=0.8\linewidth]{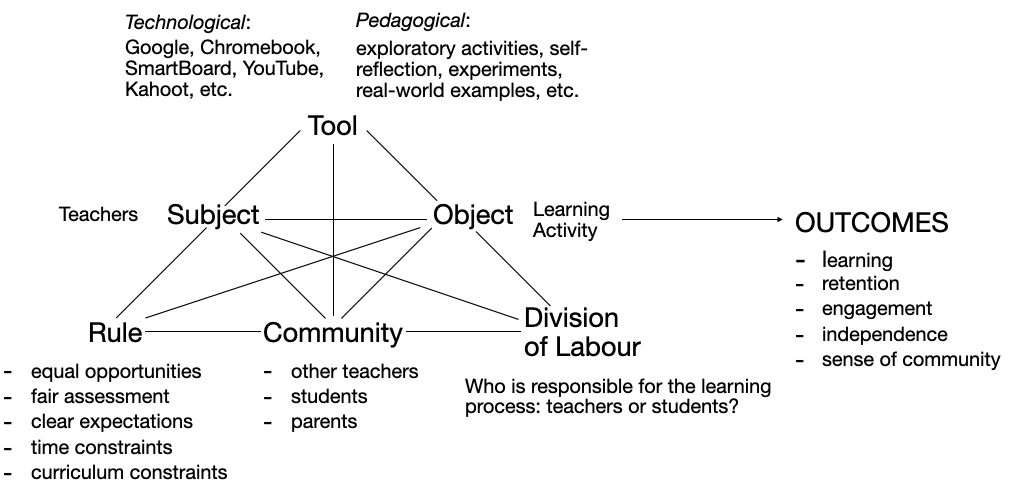}
    \caption{Derived Activity System}
    \label{fig:activitytheory_specific}
\end{figure*}

The \textbf{outcomes} that can be affected by this activity system include learning and retention, as well as students' engagement, growing independence, and sense of community. Figure \ref{fig:activitytheory_specific} shows the activity system derived from our data. The key tension points (contradictions) are discussed in detail below.  

\subsubsection{Tensions related to Division of Labour}
\label{sec:13}
A prevalent view among our teachers is that learning is a partnership between students and teachers.  As T1 said, the classroom environment ``is not just the teachers trying to get everyone to listen to what they are talking about.'' To prompt students in taking ownership of their own learning, \textit{exploratory exercises} were found to be a commonly used tool.  For example, T4 explicitly included exploration as part of the \textit{van Gogh Inspired Art} activity, where students are required to research facts that they find interesting about the artist.  Similarly, T2 takes a ``throw-them-in-the-dark'' approach for the \textit{Pulleys} activity, in which students, without being told anything about the topic, are asked to complete fill-in-the blank sheets about pulleys with a partner. T1 explained how they gave students a set of objects (e.g., a popcorn maker) and asked them to figure out how they work.  They said, ``You want the students to make those discoveries on their own...that's a huge benefit to the process.'' Sometimes, teachers encouraged student exploration simply by admitting ignorance. T1 said, ``Teachers today aren't afraid to say---I have no idea what the answer is.  Do you want to research that and let me know?'' They added, ``[Students] love to teach the teachers, so that's great too.'' 

Another form of exploration is based on the self, or `self-reflection'. This type of exploration provides a window to a student's thought process. For example, T4 asked students to provide ``thoughtful, honest reflections'', not just about their artistic creation, but the activity itself---for example, indicating which parts they enjoyed, and whether they liked the process. T4 explained, ``You don't know exactly where kids think they are.  Are you progressing?  Are you just comparing yourself to others?  Are you comparing yourself to what you think it should be?''  Further, this type of self-reflection informs how the learning activity can be adjusted to ``resonate'' more with students.  As T4 stated, ``[What] is really important to me is how they felt about the art project itself.  It also helps me gauge---would I want to do that one again? Was it successful?'' Similarly, T1 uses ``exit tickets''---a Q\&A card that students have to hand in to exit the classroom---to get students to reflect on the lesson or their learning process. For the \textit{Ideal Community} activity, T3 said, ``It was really interesting to see what they each individually came up with, and their reasoning behind it was really, really cool.'' Similarly, T2 said, ``The conversations that you hear when they are trying to figure [the worksheets] out is just genius.  It is not genius as in what they say, but how it is they are learning.'' \\

\textbf{TENSION 1: Division of Labour --- Community --- Tool.} Despite these benefits, exploration can be difficult for students because \textbf{they do not know how to navigate such a wide space of possibilities}. Research-type exploration can be challenging as the students ``don't have knowledge of what else is there outside [of what they know]" (T2). T3 observed that some students are ``completely overwhelmed'' when asked to do research related to a topic using Google search, as there is too much information out there. Learning from experience, T3 now provides students with a list of informative and trusted websites to do their research. The ability to handle exploration can also depend on students' self-confidence.  T4 mentioned, ``If they don't have the confidence to make a move without having their hand held, that's what becomes challenging'' because ``they are not fully getting involved in the creative process'', adding, ``It's just becoming too step by step, and I don't want it to be that methodical.''  Exploration of the self can be equally challenging for the same reasons. During the first few self-reflection sessions for the art project, students would ask, ``Oh, what do you mean?  What do you want me to write?" (T4).\\

\textbf{TENSION 2: Division of Labour --- Subject --- Rule.} Teachers are also \textbf{constrained by time}. While it is desirable to let students freely explore, some structure is needed in order to guide students towards the solution in a timely manner: ``...I see them once a week.  I don't have that luxury of time'' (T4). To do this, T4 stated they provide a lot of structure to their learning activity by teaching students about elements of art and providing a rubric for guidance, describing the process of exploration as ``beginning with a rough guideline and [students] exploring from there.''  In their think-pair-share activity, T1 said, ``If I give them a specific task, it's a lot more productive; otherwise, they might be just fooling around in the hallway.'' For others, having too much freedom to explore can result in time management issues.  In science fair projects, T5 sees ``right away who can manage their time and who can't.'' Some students who performed the best in directed activities, to their surprise, did not end up with a great final product due to ``the lack of organization''.  In the \textit{Pulleys} activity, T2 describes her process as a form of guided discovery---``You are slowly, slowly incorporating little questions, little blurbs that make them think.''

In short, while exploratory activities help students take ownership of the learning process, students can be paralyzed by a lack of confidence and intuition about next steps---resulting in a tension between division of labour, community, and tool.  In addition, due to the lack of time, there exists a tension between division of labour, subject, and rule, requiring structure to guide student exploration.

\subsubsection{Tensions between Community and Rule}
\label{sec:14}

\textbf{TENSION 3: Community --- Rule.} A key consideration mentioned by most teachers is \textbf{providing equal access and opportunities for students with different personalities, abilities, and needs}.  This highlights a tension between community (i.e., diverse student needs) and rule (i.e., providing equal opportunities). T1 explained, ``When I think about an activity, I want to think about all my different kinds of learners --- my auditory learners, my visual learners, my kinesthetic learners.\footnote{In the pedagogical literature, this system of classifying learners is controversial.  Here, we report teachers' perspectives as is without a critical analysis of their validity.} A great lesson is one where I can focus on a couple of different learning styles and grouping procedures.''  Another consideration, as T5 mentioned, is ``the level of the students [and] how fast the class progresses as a group.'' They explained, ``in some classes, I have to go slower, explain more, give them more hands on activities to do; some classes tend to ask more questions than the other ones.''  Especially for STEM subjects, T5 explained, ``You have to differentiate.  As you are explaining and re-explaining to the group who is struggling, you have to give something to other people who are done.''  And as T1 said, ``The hardest thing is when some of the kids are so ready, and some of them still need all that support to get there.  Within any learning environment, you have such a range of knowledge and skill.'' 

As such, teachers often have to vary the instruction to adjust to varying abilities of the students.  As T5 said, ``You do see these different personalities and how they can be differently engaged.  Some of the students, they are really interested, and then after half an hour of talking about different types of organelles and functions, you can see that they were ready to do something else.''   Teachers often have to do extra work to prepare extension activities for students.  For subjects like Math, it can be relatively easy to find more difficult exercises for the students.  But for other subjects, T1 said, ``I need to have [a] quick hitting thing that the students can go and look up..." For extension, T1 sometimes has the more advanced students ``take on a leadership role and research [another student]'s question.''  But they also cautioned that peer-to-peer teaching needs to be designed in a careful way, explaining that ``half the time'', the extension activity could be one student teaching other students, ``But I certainly don't want it to be all the time.'' Additionally, T1 mentioned that while it is beneficial for students, especially those lacking social skills, to model the teacher, ``teaching someone else is a very higher order skill'' that is difficult to master.

Further, the ability of the teacher to select groups of the right size and composition is vital to the success of a particular activity.  For example, for the \textit{Rock Identification} activity, if a group is too large, then the ``kinesthetic learners'' won't be able to ``touch [the rocks] enough'' (T1).  For T5, grouping can be beneficial for students who need that ``interaction with their friends.''  They continued, ``You can see whenever they are working in groups, or even just with a partner discussing things, they are more at ease.  They don't feel lonely any more, or they are less distracted thinking about other stuff.'' 

Lastly, a major need of students is to be able to \textbf{relate class content to what they know and what is concrete in the real world}. T5 emphasized that ``application to society and application to real life is actually part of the expectations [of the curriculum].'' Examples include mining skittles from cookies to demonstrate the difference between a rock and a mineral (T1), asking students to relate the experience of a young soldier to themselves (T3), letting students research the life of an artist (T4), or having students create their own slides to see organelles in ``real life'' under a microscope (T5). T2 said that pictures in textbooks are just not relatable: ``You don't want to do science out of a textbook, especially in a world where kids can now see the world at-the-click-of-a-button''.  T3 said, ``I often tried to make links with what's going on around the world today'', asking her students if the plight of the First Nations people reminds them of current events in the world. When teaching art, T4 said, ``some kids attach more of their learning to the life of an artist, as opposed to the art that they create; certainly, because the lives of the artists tend to be colourful, it usually piques their curiosity just from that.''  T4 explained that ``the artist becomes the springboard for new learning'' and ``makes doing art a little less daunting for kids who aren't familiar with creating art and applying new techniques.''  

In summary, there is demand for tailoring learning activities to the diverse personalities, abilities, group dynamics, and needs of the students. The tensions here relate to providing equal opportunities to students as well as following curriculum constraints.

\subsubsection{Tensions related to Subject}
\label{sec:15}

\textbf{TENSION 4: Subject --- Tool.} Learning activities allow for real-time evaluation and feedback as students' work unfolds over the course of the lesson.  As T3 said, with some students, sending work home means that, ``A lot of times, they don't do the work, mom and dad do; then, I can't see the progression. I want to be able to see so I can mark properly.''  As such, one of the strongest tensions discussed during the interviews was \textbf{the lack of efficient tools for monitoring student progress}. Currently, teachers use various tools for evaluation.  T5 has students complete ``Check Your Understanding'' sheets, some teachers use Google Classroom quizzes (T3, T5), and other teachers use oral evaluation to gauge student understanding (e.g., T2, T5).  One challenge in using questions to assess who doesn't know something is that some students who are shy are reluctant to admit in front of their peers that they are struggling.  T2 said, ``Some [students] are just quiet.  They are hiding away from you when you are asking questions or they are pretending that they are looking at something.'' T5 echoed the same sentiment: ``Some students are shy, and would keep the questions to themselves as they are afraid that it would not be a very intelligent question.  They would never ask or they would come at the end of class to ask, so their peers wouldn't hear their questions.''   \\

\textbf{TENSION 5: Subject --- Community.} The problem of monitoring is also due to a \textbf{lack of capacity}. ``You have a large class, ... sometimes they raise their hands at the same time, [and] it would be nice to have a helper so that we can address their concerns faster.  They do help each other, but sometimes, they don't have the answer'' (T5).  T2 said that having someone to watch the students would be helpful: ``I am one person, I have nineteen kids... Pulleys and gears are super safe, but if you are doing something that is moving, exploding, it could go bad in a couple of seconds.  You want someone to be there, not just you.'' Distraction is also a problem---as T1 said, especially for less hands-on activities, ``you certainly have kids all the time that need to be directed... Basically, all that amounts to is just the teacher walking around and re-directing.''

To re-iterate, the tensions related to Subject (i.e., the teachers) point to the fact that teachers currently lack the tools and capacity to efficiently evaluate and monitor student progress as they complete the activity.

\subsection{\textbf{RQ2 \& RQ3: How do teachers envision a social robot being incorporated and how might it alleviate or exacerbate tensions?}}
\label{sec:16}

The teachers envisioned various functions and behaviours a social robot could perform to support learning activities.  Some of the ideas came from the teachers' current strategies for dealing with the tensions; others were discovered as a result of teachers puppeteering the robot. We present here the ways teachers felt the social robot could be used, the verbal behaviours suggested, and how they relate to the current tensions in the activity system.  The learning activities of all teachers involved both lecture-style teaching as well as group- or individual-work. We distinguish between teacher vs. no teacher involvement, as during the puppeteering, it became clear that there were differences in how teachers envisioned using the social robot in the two scenarios.

\subsubsection{Lecture/Discussion: Teacher-Robot Interaction}
\label{sec:17}
During lecture-style, in which the teacher is addressing the class as a whole, teachers felt that they could use the robot to teach the material and increase engagement with students, in two main ways:\\

\textbf{Teacher asking robot.} Most frequently the robot was puppeteered as a peer to the students, that the teacher could ask questions. T1, when role-playing as the teacher, encouraged the robot to answer by saying things like, \textit{``Don't be shy!"} and \textit{``I bet NAO knows!"}. They explained, they wanted to ``have [the robot] do the teaching part of it. I think the kids would rather, rather listen to his explanation than mine." T2 prompted playful back-and-forth exchanges with the robot while role-playing as the teacher, e.g., \textit{Teacher: ``What do robots have?" Robot: ``I have motors!" Teacher: ``You do?" Robot: ``Yeah!" Teacher: ``Oh, well congratulations, that's fantastic that you have motors!"} When the robot replied to questions, T2 mentioned that they would like the robot to give both correct and incorrect answers --- so that students could correct it, ``because they learn through that struggle and correcting others." T4 mentioned that some questions are tough for students to answer, resulting in a low number of responses---providing a perfect opportunity for the robot to step in. When answering teacher questions, T5 had the robot express uncertainty, \textit{``I am not sure"}, as well as hypothesizing, \textit{``I think...."}. T1 said, ``It would be great, to have [the robot] in the position of being the expert, yet sometimes saying, `I don't get it'.''  T1 sees this as an objective for students: ``That's the way we want them to feel all the time, that everyone has that expertise that they are going to offer at some point...that when I do feel comfortable I share, and when I don't know I ask questions.'' Having the robot be both a teacher and a learner simultaneously can guide students more quickly toward solutions (Tension 2). It also models desirable student behaviour and could aid in promoting their self-confidence (Tension 1). 

A major theme in Tension 3 is that in order to reach students, teachers need to make the materials relatable.  In the same vein, teachers think that the robot, both in terms of what it should say and how it should say it, should be relatable to students.  T2 said, ``[Students] understand when they [can] relate [the lesson] to their life'' and ``usually they learn so much better from peers because they are talking in Grade 4 language. That's what they relate to.''  In puppeteering, T1 and T2 (both currently Grade 4 teachers) used their current students as a point of reference for how the robot should speak.  T2 said that they had the robot ``use the same logic that kids would'', such as ``blurting things as fast as they can without thinking about it.'' \\

\textbf{Robot asking teacher.} In other instances, teachers felt the robot could be the one asking the questions, e.g., ``I think I could see [the robot] more in the audience, just asking some good questions, that maybe the kids haven’t thought about..." (T4). They elaborated that, usually they were the ones asking questions of the students, but they would like to have the robot asking ``the how and the why". T3 felt that because the students do a lot of learning through conversation, they would like the robot to be able to learn from the conversations happening around it and then contribute to discussions using what its learned. By having the robot ask questions in front of the class, the robot can model or prompt exploration (Tension 1), the teachers can use the robot to guide students (Tension 2), and the robot can ask questions students may be too shy to ask (Tension 3).

However, T3 mentioned that the robot might be useful for engaging students who are more reserved in small group settings (Tensions 3 \& 5), but in a large-group ``I think it would probably be a distraction to some of the kids, and it would take my time, or my attention away from them.'' Similarly, T3 pointed out that the robot's presence could make some students more reluctant to speak up: ``Some of the students are a bit harder to engage than others, some of them are very very very shy, very shy. ... So in a big classroom some kids might feel silly to be talking to a robot.  I don't even want to raise my hand when I'm talking with people let alone [a robot], right?". This could introduce a new Tension between Community and Object if students become less comfortable and engaged during the learning activity.\\ 

In both question-asking scenarios, teachers frequently had the robot exhibit a playful tone: \textit{``No way!"}, express engagement and interest: \textit{``Really?"}, \textit{``Cool!"}, \textit{``Aah"}, as well as use humour:\textit{ Student: ``Where did you hear that?" Robot: ``From a documentary I watched last week!"}. Additionally, T1 had the robot tell various jokes to the class, tapping into students' love for puns to help reinforce learning.

\subsubsection{Group/Individual: Student-Robot Interaction}
\label{sec:18}
During the group (two or more students)/individual (one student) scenario, the teachers had the robot exhibit a number of different verbal behaviours when they role-played. The teachers commonly envisioned the robot sitting at a table with one or more students in a group, and interacting with them to achieve the following functions:\\

\textbf{Providing Guidance.} Teachers puppeteered the robot to use both questions and statements to \textbf{direct}, \textbf{reinforce}, and \textbf{correct} students. They had the robot ask for explanations, e.g., \textit{``What does intrusive mean?''}, confirm students' ideas, e.g., \textit{``Layers is good''} or \textit{``Change is a great word because...''}, offer their own answers, e.g., \textit{``I think it means inside''}, and correct students if they misunderstood, e.g., \textit{``It is an example, not what the word means.''} T1 explained, ``like a student [the robot] is offering his input, and I think he's furthering, like clarifying, which I think is really good ... [The robot] is adding a bit of direction to the learning, to moving the learning process on.'' T3 mentioned that ``kids end up really on the wrong track, way off topic'', and if ``they are not really understanding the activity, or if they are missing some information, the robot can certainly jump in, and refocus their effort''. Similarly, as T4 said, the robot can ``reinforce the expectations of the assignments'' and remind the students about the learning activity outcomes by asking questions.
T2 envisioned that the robot can make the students ``think about whether their answer is right, or take them one step further and say---`why do you think that?'". With these behaviours, a social robot can be used to guide students by providing structure to learning activities (Tension 2) and keeping students focused and on the right track (Tension 5).\\

\textbf{Promote Learning by Teaching.} T5 envisioned that the robot could ask questions like, ``Can you please help me remember what an organelle is?'', providing the right answer when students are unsure. Similarly, T2 proposed ``Maybe if they hear [the robot] saying wrong answers, they question it, or if [the robot] were to say `Why?', then they'd explain it.'' Having the robot withhold information from time to time can help encourage students to question the robot's authority, become more independent learners, and teaching a robot could help build self-confidence (Tension 1)---``They would feel like they are so cool, that they know better than a robot.  That's a confidence boost'' (T2). Furthermore, by asking students to teach, the robot can enable students with different levels of abilities to be equally challenged and provide an extension activity to those students who are ahead (Tension 3). In this capacity it can also serve as a form of assessment or monitoring of the students' knowledge (Tension 4). As T2 mentioned,  ``If [students] are confused, they might listen to other students' explanations...I want to hear that conversation...''. 

Although learning by teaching can be beneficial for students, teachers do not want students to take on the role of a teacher too often as they could be mislead if the robot makes mistakes and be taken further away from the target objectives of the activity, or become too focused on the robot's learning rather than their own. These events could exacerbate Tension 2.\\

\textbf{Relating/Sharing Knowledge.} Teachers also used the robot to interject facts during the task. They would do this by having the robot say for example, \textit{``Did you know that...''} or when conversing about the rock type granite, T1 had the robot say, \textit{``My kitchen has granite counter tops”}, as a way of linking the material directly to students' home life. T5 stated, ``Oh that would be great if the robot can tell them something from its own experience.'' They also mentioned that students ``love to talk”, and ``a lot of time they share some of their experiences” relating the lesson (e.g., mixture separation) to the real world (e.g., scooping fat off the soup), so maybe ``they can tell something to the robot and then the robot can tell them, `Oh, I see how that relates to what we've discussed’.” In this way, the robot can be an \textbf{information provider} (Tensions 3 \& 5). 

However, the robot should not ``dominate the process of the conversation'' (T1), as if the robot is too much of a leader or know-it-all, that could discourage students from taking ownership (exacerbating Tension 1). T2 cautioned, ``I wouldn't find it useful if [the robot] would just blurt out correct information.  I would want [the robot] to have them question it.'' Similarly, T3 mentioned that they would want the robot to be ``truly interactive'' and contribute to discussions --- ``it would be more valuable than having the robot... regurgitating facts". \\

\textbf{Offer Emotional Support.} To facilitate an empathizing or supportive function, teachers puppeteered the robot in a number of ways. In response to a student's boredom, T2 had the robot mirror their sentiment. They explained that a lot of students will say, `I'm bored', because they don't know the answer and don't want to admit it. As they explained, ``that's what I was doing for this [robot] too, because at the end of the day, the students need to figure it out on their own.''
Others used statements of agreement with the student, \textit{``I think so too!''}, and asking students for their opinion, \textit{``What do you think?"}. As T4 suggested, ``[The robot] could prompt kids to ask those high-level thinking questions to see if they are really understanding the project itself.'' In the puppeteering session, T4 had the robot say, \textit{``I really like the horizon line. Do you know anything about perspective?''}.  When asked why they had the robot pose that question, T4 explained, ``I wanted to see if you were understanding, or intentionally, using [that element of art].''  By prompting self-reflection, the robot can help foster independence (Tension 1) and monitor student progress despite the potential discrepancy in learning styles and speed (Tensions 3 \& 4).

T4 envisioned that the mere presence of a robot that the students can converse with can be a \textit{good distraction}, because: ``You are not thinking so much about what you are putting on the page... maybe it lightens the mood a little.''  When T4 puppeteered the robot, they had it act as a not-so-confident peer and encouraged the student to provide it with encouragement: e.g., \textit{Robot: ``You asked me if I can draw.  I'm terrible at drawing.''  Student: ``Anyone can draw.  Our teacher said so!'' Robot:``You're right.  Like everything, it takes practice!''}. The robot can also act as a middleman, a companion students can approach if they are hesitant to approach the teacher or their peers.  For example, the robot can ask a shy student to write one question on a flashcard, and give it back to the robot, who would then read it to the class.  The robot can individually ask students if they are concerned about a topic, a person, or anything in general.  The robot could help answer questions that students are ``too embarrassed to ask their friends'' (T5).  

All teachers had the robot use words of encouragement, e.g., \textit{``We are awesome!'}, \textit{``Cotton balls! I love it!''}. In one activity, students are given mixtures to separate without being explicitly told what to do.  For this, T5 imagined that the robot could give hints and encourage students to experiment.  As T5 explained, ``A lot of times, students are reluctant to try new things because they are afraid it's going to mess it up...the robot could be the kid that wants to try lots of things---Let's use this!  Let's try this and see what happens!  Sometimes he gets it right, and sometimes he gets it wrong.'' They added, ``There are kids that wanna try everything, and that's a struggle. The robot could say `Stop doing this! Let's do it step by step'.  So, the robot could be like one of those eager kids but not too reckless.''

In short, by providing \textbf{emotional support}, the robot can reach those students who are more reserved or struggling and act as a helper to the teacher, thus alleviating aspects of Tensions 3, 4, and 5. Furthermore, the robot can help those students lacking self-confidence and initiative to explore (Tension 1) by directly encouraging them with hints or by modeling the desired student behaviour by being confident. One caveat however, is that self-reflection requires a certain level of maturity. Age may be a factor---for example, T2 and T3 expressed concerns about the students' ability to reflect on their sense of self, while T4 who teaches older students expects that they are capable of meaningfully reflecting on their product and process.

\subsubsection{Other Scenarios and Behaviours}
\label{sec:19}
Although lecture-style and group-work were the only scenarios role-played, there were other types of learning activities that the teachers felt a social robot could be useful in. For example, T3 mentioned, ``I think I really see the value of presentations for something like this. And it would also engage the kids who might not otherwise pay attention." One specific idea is the robot as a `surrogate' presenter, who learns a topic from a student and then presents in front of the class on behalf of or alongside that student.  Team teaching ideas also emerged during the brainstorming session. Teachers can envision multiple robots being taught by different groups of students and then competing with each other on who learned the most. 

Other possible interactions mentioned by teachers included: the robot playing music to encourage and praise the students for their hard work, students could make skits with the robot as one of the actors representing a certain part of the lesson, students and robot could play established games such as, `Vibrant Verbs' or `Amazing Adjectives', the robot can set goals for the class by providing comparison data daily, e.g., class participation was highest this week on Wednesday when we discussed `x', the robot could do ``a dance of shame'' should a student get three items wrong during an activity, or a spontaneous happy dance to surprise students who get three items correct. More generally, the robot and student can celebrate a job well done together by dancing, or with something simpler like a high five. These ideas could alleviate certain aspects of Tensions 1 and 5, and support a role for the robot as a playful learner, and a ``funny'' personality rather than an overly serious one.

\section{Discussion}
\label{sec:20}

As teachers are principal stakeholders in the use and application of educational technologies \cite{nordkvelle2005visions}, it is essential to understand how they feel social robots can be used in the classroom and what pain points in their current procedures it can alleviate or exacerbate. 
Previous investigations of teachers' perspectives of classroom robots have focused on one topic (e.g., language learning \cite{chang10}), activity (e.g., a map-reading task \cite{serholt2014teachers}), role (e.g., robot as teacher \cite{newton2019humanoid}), or behaviour (e.g., adaptive \cite{Ahmad2016}). Our approach, in contrast and supplement to prior work, draws information from a small number of teachers discussing a {\it diverse} range of learning activities, across various age groups (Grades 4-8), domains (visual arts, science, technology, history, geography), formats (lecture, hands-on activities, worksheets, discussions), and styles (directed, exploratory).

Framing the context surrounding learning activities using activity theory illustrated that the teachers currently face a number of tensions. Two are related to \textbf{division of labour}, i.e., who is responsible for the learning process: teachers or students? The teachers predominantly felt that learning is a partnership between students and teachers, and to prompt students to take ownership of their own learning they employed exploratory exercises. However, a lack of confidence and intuition on how to use available resources (tool) when asked to explore can hinder students (community), and although it is desirable to allow students to explore freely, teachers (subject) are constrained by time (rule) and require structure to guide students in a timely manner. Another two are associated with \textbf{community}, i.e., diverse student needs. That is, teachers often have to vary instruction to provide equal access and opportunities (rule), and a curriculum constraint (rule) dictates a need for students to be able to relate class content to the real world. Lastly are tensions pertaining to \textbf{subject} - the teachers - and aspects of assessment and evaluation. The teachers discussed various tools they currently use for evaluation, however were clear that they lack efficient means (tool) and capacity (community) for monitoring student progress.

The teachers described their current use of technology to get students' attention (T1, T3), reinforce and make vivid abstract concepts (T2, T5), as an extension tool for students who have finished their assigned work (T1), as a review tool (T2, T3), and as a way to help students ``relax and enjoy themselves'' (T3). Overall, the teachers responded positively to the idea of incorporating a social robot as a technological tool to support learning activities. The manners in which they envisioned the use of a social robot during learning activities in some ways mirror how they currently use technology, but it provides them with a way to alleviate certain tensions.

\paragraph{Teacher-Robot Interaction} First, unlike prior work showing teachers envisioning the preferred use of a tutor robot in group-settings \cite{serholt2014teachers}, most of our teachers could see the benefit in scenarios both with and without their involvement---albeit in different ways. While addressing the class, teachers envisioned the robot both asking questions of the teacher, as well as responding to teacher questions. Through this interaction they feel the robot can \textbf{model desirable student behaviour}, \textbf{engage students through conversation}, and \textbf{make the material relatable} not only with \textit{what} it says but \textit{how} it says it, i.e., for Grade 4, the robot's language can be based around Grade 4 student language. This type of functionality could alleviate tensions associated with division of labour (e.g., providing structure to guide students because of a lack of time), community (adhering to the curriculum), and subject (e.g., a way of monitoring student progress --- if the robot answers questions incorrectly, students can correct it and the teacher can be a part of this process).

\paragraph{Student-Robot Interaction} When students are working on a learning activity without teacher involvement (i.e., in a group or alone), teachers envisioned the robot for four high-level functions: \textbf{guiding}, \textbf{promoting learning by teaching}, \textbf{sharing knowledge}, and \textbf{offering emotional support}. Guidance can be achieved through enforcing expectations by having the robot ask students directly what they would do differently next time, or what would be a logical next step in the activity.  In a more indirect fashion, the robot can direct, reinforce, and correct students by asking them to explain parts of the task, acknowledge and confirm student ideas or offer up their own, and correct students if they misunderstand or are going off topic. These behaviours can provide structure to the activity (division of labour) and a means for monitoring student progress (tool).
In order to promote learning by teaching and increase students' confidence, just as teachers can feign ignorance to encourage exploration, the robot can act as a novice or less knowledgeable peer and ask students for help. This behaviour can be used during learning activities or as extension activities needed for those students that are ahead - mitigating aspects of tensions associated with community and rule (i.e., providing equal opportunities to students with diverse needs). By having the robot share knowledge, it can relate the learning material to the real-world using its own experiences --- a requirement in the curriculum (rule). 
Providing emotional support to students was another commonly expressed function. Teachers accomplished this by having the robot mirror students' sentiments, agreeing with them, asking for their opinions, prompting them to self-reflect, and offering words of encouragement. In this way, the robot can alleviate aspects of tensions related to community (the students' needs) and subject (teachers lacking efficient tools and capacity for monitoring).\\

Although the teachers responded positively to the idea of using a social robot for learning activities, they recognized potential costs as well. Similar to prior work \cite{Westlund2016}, distraction is a concern for some teachers (T1, T2, T3) --- possibly exacerbating tensions around division of labour (i.e., lack of time) and rule (i.e., providing equal opportunities) as teachers are already constrained by a lack of time, if some students are distracted it could take teachers' time and attention away from the rest of the class. Distraction could also introduce tensions related to the object (the learning activity) and the outcomes, i.e., if students feel uncomfortable and less engaged in the learning activity.  Many teachers mentioned that while students will be fascinated by the robot, they may also misbehave in its presence, and the robot would have to respond, for example, by saying ``Stop teasing me!'', a concern similarly expressed by teachers in Diep, Cabibihan, and Wolbring's \cite{diep2015} study. Further, there was apprehension related to failures in the technology exacerbating the lack of time constraint related to division of labour: ``I don’t know about like class-wide ... if there's like time delays between like the robot and the discussion, or if there's pronunciation [issues]..." (T3). Finally, given that social robots can be expensive to purchase, teachers expressed worry about the ratio of students to robot, and how to form groups such that particular students, e.g., those who are more technically savvy, will dominate and monopolize the robot - exacerbating the existing demand between community and rule, for tailoring learning activities to group dynamics and student needs.  Similar concerns were raised by the teachers in Serholt et al.'s investigation \cite{serholt2014teachers}.

Reviews on social robots for education \cite{belpaeme2018social,mubin13} report that robots commonly take on the role of a teacher, peer, or novice. Certain functions may be more amenable to certain robot roles --- for example, a {\it teacher} robot can enforce expectations and prompt self-reflection, while the {\it novice} robot can feign ignorance.  However, in practice, our results indicate that the view that roles and functions map one-to-one is too simplistic.  For example, a {\it teacher} robot that asks questions and admits not knowing something, may in fact be more beneficial to students, as they see that experts are learners as well.  Depending on the goals of a learning activity, the composition of the class, and learning styles, abilities, and progress of individual students, the robot can adopt multiple roles as well as different roles over time.

Our work supports previous findings on teachers' perspectives reporting similar suggestions, i.e., the robot being a motivator, providing feedback, and encouraging students (e.g., \cite{serholt2014teachers,Ahmad2016}), and extends it to include specific behaviours teachers envision to support these functions. It also shows differences in usage with and without teacher involvement (adding to our understanding of social robot behaviours for supporting one-on-one, two-persons to a robot, and small group interactions, e.g., \cite{rosenberg2020}), and how introduction of a social robot can affect the context surrounding learning activities. This provides suggestions for improving the design of educational social robots for learning activities in primary and middle school classrooms. And as Broadbent et al.'s \cite{Broadbent2018} findings suggest, students in this age range (i.e., around 5-12 years) and their teachers, are particularly positive about the usefulness of social robots in their classrooms, and therefore stakeholder feedback is of particular importance for the improvement of such technologies.

\subsection{Limitations}
\label{sec:21}
We acknowledge a number of limitations with this work. It was completed at one school (Kindergarten to Grade 8), with a small sample size, and the discussions were based around North American school practices. Additionally, the teachers had little experience with (social) robots, were introduced to only one kind of robot (anthropomorphic), with one speaking and gesture-style shown, which may have influenced the results. Although a small sample, prior work investigating teachers' perspectives has been published with similarly small (less than 10) numbers of participants (e.g., \cite{serholt2014teachers,Ahmad2016,chang10}) and less time with each participant (e.g., 30 minutes \cite{Ahmad2016}). We emphasize that our approach is not aimed at generalizing across populations, but rather at contributing a rich description of the socio-technical situation under investigation.

\section{Conclusion and Future Directions}
\label{sec:22}
Through interviews, role-playing with robot puppeteering, and group brainstorming, we explore how teachers envision the use of a social robot for learning activities. In contrast to prior work, our approach goes beyond the coarse-grained classification of robots as a teacher, peer, or novice, to finer-grained functionalities and behaviours that define {\it how the robot should act} in different situations and \textit{why}. Using activity theory, we frame learning activities as a system of interacting components to understand the tensions that exist, and describe concrete ways teachers envision the robot alleviating these tensions in learning activities. In future work, we will implement these functional roles as a modular system and conduct studies to elucidate the gap between teachers' expectations and reality.

\section*{Compliance with Ethical Standards}
\textit{Conflicts} The authors declare that they have no conflict of interest.\\
\textit{Ethics Approval} The study obtained ethics clearance from the University of Waterloo Research Ethics Committee \newline (ORE \#40392).\\
\textit{Informed Consent} Informed consent was obtained from all participants recruited for the study.\\
\textit{Funding} This work was made possible by funding from \newline NSERC Discovery Grant RGPIN-2015-0454.

\bibliographystyle{spbasic}      
\bibliography{references.bib}

\end{document}